\documentclass[twoside,11pt]{article}

\usepackage{arxiv}

\usepackage{authblk}
\usepackage{amsmath,amsfonts,amssymb}
\usepackage{bm}
\usepackage{xcolor}
\usepackage{booktabs}
\usepackage{tikz}
\usetikzlibrary{shapes,arrows,positioning,fit,calc}
\tikzset{block/.style={draw, thick, text width=3cm, minimum height=1.5cm, align=center},line/.style={-latex}} 
\usepackage{algorithm}
\usepackage{algorithmicx}
\usepackage{caption}
\usepackage[colorlinks=true,linkcolor=blue,citecolor=blue]{hyperref}
\usepackage[sort, numbers]{natbib}

\begin{document}

\title{LINFA: a Python library for variational inference with normalizing flow and annealing}

\author[]{Yu Wang}
\author[]{Emma R. Cobian}
\author[]{Jubilee Lee}
\author[]{Fang Liu}
\author[]{Jonathan D. Hauenstein}
\author[]{Daniele E. Schiavazzi}
\affil[]{Department of Applied and Computational Mathematics and Statistics\\ University of Notre Dame, Notre Dame, IN, USA}
\date{ }

\maketitle

\begin{abstract}
 Variational inference is an increasingly popular method in statistics and machine learning for approximating probability distributions.
We developed LINFA (Library for Inference with Normalizing Flow and Annealing), a Python library for variational inference to accommodate computationally expensive models and difficult-to-sample distributions with dependent parameters. 
We discuss the theoretical background, capabilities, and performance of LINFA in various benchmarks. LINFA is publicly available on GitHub at \url{https://github.com/desResLab/LINFA}.
\end{abstract}


\section{Introduction}

Generating samples from a posterior distribution is a fundamental task in Bayesian inference.
The development of sampling-based algorithms from the Markov chain Monte Carlo family~\cite{metropolis1953equation, hastings1970monte, geman1984stochastic, gelfand1990sampling} has made solving Bayesian inverse problems accessible to a wide audience of both researchers and practitioners.
However, the number of samples required by these approaches is typically significant and the convergence of Markov chains to their stationary distribution can be slow especially in high-dimensions. Additionally, satisfactory convergence may not be always easy to quantify, even if a number of metrics have been proposed in the literature over the years.
More recent paradigms have been proposed in the context of variational inference~\cite{wainwright2008graphical}, where an optimization problem is formulated to determine the optimal member of a parametric family of distributions that can approximate a target posterior density.
In addition, flexible approaches to parametrize variational distributions through a composition of transformations (closely related to the concept of \emph{trasport maps}, see, e.g.,~\cite{villani2009optimal}) have reached popularity under the name of \emph{normalizing flows}~\citep{rezende2015variational,dinh2016density,kingma2016improved,kobyzev2020normalizing,papamakarios2021normalizing}.
The combination of variational inference and normalizing flow has received significant recent interest in the context of general algorithms for solving inverse problems~\cite{rezende2015variational}.

However, cases where the computational cost of evaluating the underlying probability distribution is significant occur quite often in engineering and applied sciences, for example when such evaluation requires the solution of an ordinary or partial differential equation. 
In such cases, inference can easily become intractable. Additionally, strong and nonlinear dependence between model parameters may results in difficult-to-sample posterior distributions characterized by features at multiple scales or by multiple modes.
%
The LINFA library is specifically designed for cases where the model evaluation is computationally expensive. In such cases, the construction of an adaptively trained surrogate model is key to reducing the computational cost of inference~\cite{wang2022variational}. 
In addition, LINFA provides an adaptive annealing scheduler, where temperature increments are automatically determined based on the available variational approximant of the posterior distribution. Thus, adaptive annealing makes it easier to sample from complicated densities~\cite{cobian2023adaann}.

This paper is organized as follows. The main features of the LINFA library are discussed in Section ~\ref{sec:capabilities}, followed by a brief outline of a few selected numerical tests in Section~\ref{sec:benchmarks}. Conclusions and future work are finally discussed in Section~\ref{sec:conclusions}. The paper is completed by a brief description of the background theory and reference to the relevant papers in Appendix~\ref{sec:background}, a detailed presentation of a four benchmarks in Appendix~\ref{sec:detailed_benchmarks}, and a list of all the relevant hyperparameters in Appendix~\ref{sec:hyper}.

\section{Capabilities}\label{sec:capabilities}

LINFA is designed as a general inference engine and allows the user to define custom input transformations, computational models, surrogates, and likelihood functions.

\begin{enumerate}
\item[{\bf 1 -}] {\bf User-defined input parameter transformations} - Input transformations may reduce the complexity of inference and surrogate model construction in situations where the ranges of the input variables differ substantially or when the input parameters are bounded. A number of pre-defined univariate transformations are provided, i.e, \emph{\texttt{identity}}, \emph{\texttt{tanh}}, \emph{\texttt{linear}}, and \emph{\texttt{exp}}. These transformations are independently defined for each input variable, using four parameters $(a,b,c,d)$, providing a nonlinear transformation between the \emph{normalized} interval $[a,b]$ and the \emph{physical} interval $[c,d]$. Additional transformations can be defined by implementing the following member functions.
\begin{itemize}
\item \emph{\texttt{forward}} - It evaluates the transformation from the normalized to the physical space. One transformation needs to be defined for each input. For example, the list of lists 
\begin{verbatim}
trsf_info = [['tanh',-7.0,7.0,100.0,1500.0],
             ['tanh',-7.0,7.0,100.0,1500.0],
             ['exp',-7.0,7.0,1.0e-5,1.0e-2]]
\end{verbatim}
defines a hyperbolic tangent transformation for the first two variables and an exponential transformation for the third. 
\item \emph{\texttt{compute\_log\_jacob\_func}} - This is the log Jacobian of the transformation that needs to be included in the computation of the log posterior density to account for the additional change in volume.
\end{itemize}
\item[{\bf 2 -}] {\bf User-defined computational models} - LINFA can accommodate any type of models from analytically defined posteriors with the gradient computed through automatic differentiation to legacy computational solvers for which the solution gradient is not available nor easy to compute. New models are created by implementing the methods below.
\vspace{-3pt}
\begin{itemize}\itemsep -3pt
\item \emph{\texttt{genDataFile}} - This is a pre-processing function used to generate synthetic observations. It computes the model output corresponding to the default parameter values (usually defined as part of the model) and adds noise with a user-specified distribution. Observations will be stored in a file and are typically assigned to \texttt{model.data} so they are available for computing the log posterior.
\item \emph{\texttt{solve\_t}} - This function solves the model for multiple values of the \emph{physical} input parameters specified in a matrix format (with one sample for each row and one column for each input parameter dimension).
\end{itemize}
\item[{\bf 3 -}] {\bf User-defined surrogate models} - For computational models that are too expensive for online inference, LINFA provides functionalities to create, train, and fine-tune a \emph{surrogate model}. The \emph{\texttt{Surrogate}} class implements the following functionalities: 
\vspace{-3pt}
\begin{itemize}\itemsep -3pt
\item A new surrogate model can be created using the \emph{\texttt{Surrogate}} constructor. 

\item The \emph{\texttt{limits}} (i.e. upper and lower bounds) are stored as a list of lists using the format \emph{\texttt{[[low\_0, high\_0], [low\_1, high\_1], ...]}}.

\item A \emph{pre-grid} is defined as an a priori selected point cloud created inside the hyper-rectangle defined by \emph{\texttt{limits}}. The pre-grid can be either of type \emph{\texttt{'tensor'}} (tensor product grid) where the grid order (number of points in each dimension) is defined through the argument \emph{\texttt{gridnum}}, or of type \emph{\texttt{'sobol'}}, in which case the variable \emph{\texttt{gridnum}} defines the total number of samples.

\item Surrogate model Input/Output. The two functions \emph{\texttt{surrogate\_save()}} and \emph{\texttt{surrogate\_load()}} are provided to save a snapshot of a given surrogate or to read it from a file. 
\item The \emph{\texttt{pre\_train()}} function is provided to perform an initial training of the surrogate model on the pre-grid. In addition, the \emph{\texttt{update()}} function is also available to re-train the model once additional training examples are available. 
\item The \emph{\texttt{forward()}} function evaluates the surrogate model at multiple input realizations. If a transformation is defined, the surrogate should always be specified in the \emph{normalized domain} with limits coincident with the normalized intervals.
\end{itemize}

\item[{\bf 4 -}] {\bf User-defined likelihood} - A user-defined likelihood function can be defined by passing the parameters, the model, the surrogate and a coordinate transformation using \emph{\texttt{log\_density(x, model, surr, transf)}}
and then assigning it as a member function of the \emph{\texttt{experiment}} class using:

\vspace{3pt}

\emph{\texttt{exp.model\_logdensity = lambda x: log\_density(x, model, surr, transf)}}.
%

\item[{\bf 5 -}] {\bf Linear and adaptive annealing schedulers} - LINFA provides two annealing schedulers by default. The first is the \emph{\texttt{'Linear'}} scheduler with constant increments. The second is the \emph{\texttt{'AdaAnn'}} adaptive scheduler 
\cite{cobian2023adaann} with hyperparameters reported in Table~\ref{tab:adaann}. For the AdaAnn scheduler, the user can also specify a different number of parameter updates to be performed at the initial temperature $t_{0}$, final temperature $t_{1}$, and for any temperature $t_{0}<t<1$. Finally, the batch size (number of samples used to evaluate the expectations in the loss function) can also be differentiated for $t=1$ and $t<1$. 

\item[{\bf 6 -}] {\bf User-defined hyperparameters} - A complete list of hyperparameters with a description of their functionality can be found in Appendix~\ref{sec:hyper}.
\end{enumerate}

\section{Numerical benchmarks}\label{sec:benchmarks}

We tested LINFA on multiple problems. These include inference on unimodal and multi-modal posterior distributions specified in closed form, ordinary differential models and dynamical systems with gradients directly computed through automatic differentiation in PyTorch, identifiable and non-identifiable physics-based models with fixed and adaptive surrogates, and high-dimensional statistical models. Some of the above tests are included with the library and systematically tested when pushing the master branch on GitHub. A detailed discussion of these test cases is provided in Appendix~\ref{sec:detailed_benchmarks}. LINFA can be installed through the Python Package Index (PyPI) typing

\vspace{3pt}

\emph{\texttt{pip install linfa-vi}}

\vspace{3pt}

To run the tests type

\vspace{3pt}

\emph{\texttt{python -m unittest linfa.linfa\_test\_suite.NAME\_example}}

\vspace{3pt}

where \emph{\texttt{NAME}} is the name of the test case, either \emph{\texttt{trivial}}, \emph{\texttt{highdim}}, \emph{\texttt{rc}}, \emph{\texttt{rcr}}, or \emph{\texttt{adaann}}.

\section{Conclusion and Future Work}\label{sec:conclusions}

In this paper, we have introduced the LINFA library for variational inference, briefly discussed the relevant background, its capabilities, and report its performance on a number of test cases. Some interesting directions for future work are mentioned below.

Future versions will support user-defined privacy-preserving synthetic data generation and variational inference through differentially private gradient descent algorithms. This will allow the user to perform inference tasks while preserving a pre-defined privacy budget, as discussed in ~\cite{su2023differentially}.
LINFA will also be extended to handle multiple models. This will open new possibilities to solve inverse problems combining variational inference and multi-fidelity surrogates~\cite[see, e.g.,][]{siahkoohi2021preconditioned}.
%
In addition, for inverse problems with significant dependence among the parameters, it is often possible to simplify the inference task by operating on manifolds of reduced dimensionality~\cite{brennan2020greedy}. New modules for dimensionality reduction will be developed and integrated with the LINFA library.
Finally, the ELBO loss typically used in variational inference has known limitations, some of which are related to its close connection with the KL divergence. Future versions of LINFA will provide the option to use alternative losses.

\section*{Acknowledgements}

The authors gratefully acknowledge the support from the NSF Big Data Science \& Engineering grant \#1918692 and the computational resources provided through the Center for Research Computing at the University of Notre Dame. DES also acknowledges support from NSF CAREER grant \#1942662.

\appendix

\section{Background theory}\label{sec:background}

\subsection{Variational inference with normalizing flow}
Consider the problem of estimating (in a Bayesian sense) the parameters $\bm{z}\in\bm{\mathcal{Z}}$ of a physics-based or statistical model
\[
\bm{x} = \bm{f}(\bm{z}) + \bm{\varepsilon},
\]
from the observations $\bm{x}\in\bm{\mathcal{X}}$ and a known statistical characterization of the error $\bm{\varepsilon}$.
We tackle this problem with variational inference and normalizing flow. A normalizing flow (NF) is a nonlinear transformation $F:\mathbb{R}^{d}\times \bm{\Lambda} \to \mathbb{R}^{d}$ designed to map an easy-to-sample \emph{base} distribution $q_{0}(\bm{z}_{0})$ into a close approximation $q_{K}(\bm{z}_{K})$ of a desired target posterior density $p(\bm{z}|\bm{x})$.
This transformation can be determined by composing $K$ bijections 
\begin{equation}
\bm{z}_{K} = F(\bm{z}_{0}) = F_{K} \circ F_{K-1} \circ \cdots \circ F_{k} \circ \cdots \circ F_{1}(\bm{z}_{0}),
\end{equation}
and evaluating the transformed density through the change of variable formula~\cite[see][]{villani2009optimal}.

In the context of variational inference, we seek to determine an \emph{optimal} set of parameters $\bm{\lambda}\in\bm{\Lambda}$ so that $q_{K}(\bm{z}_{K})\approx p(\bm{z}|\bm{x})$. Given observations $\bm{x}\in\mathcal{\bm{X}}$, a likelihood function $l_{\bm{z}}(\bm{x})$ (informed by the distribution of the error $\bm{\varepsilon}$) and prior $p(\bm{z})$, a NF-based approximation $q_K(\bm{z})$ of the posterior distribution $p(\bm{z}|\bm{x})$ can be computed by maximizing the lower bound to the log marginal likelihood $\log p(\bm{x})$ (the so-called \emph{evidence lower bound} or ELBO), or, equivalently, by minimizing a \emph{free energy bound}~\cite[see, e.g.,][]{rezende2015variational}.
\begin{equation}\label{equ:ELBO}
\begin{split}
\mathcal{F}(\bm x)& = \mathbb{E}_{q_K(\bm z_K)}\left[\log q_K(\bm z_K) - \log p(\bm x, \bm z_K)\right]\\
& = \mathbb{E}_{q_0(\bm z_0)}[\log q_0(\bm z_0)] - \mathbb{E}_{q_0(\bm z_0)}[\log p(\bm x, \bm z_K)] - \mathbb{E}_{q_0(\bm z_0)}\left[\sum_{k=1}^K \log \left|\det \frac{\partial \bm z_k}{\partial \bm z_{k-1}}\right|\right].
\end{split}
\end{equation}

For computational convenience, normalizing flow transformations are selected to be easily invertible and their Jacobian determinant can be computed with a cost that grows linearly with the problem dimensionality. 
Approaches in the literature include RealNVP~\cite{dinh2016density}, GLOW~\cite{kingma2018glow}, and autoregressive transformations such as MAF~\cite{papamakarios2018masked} and IAF~\cite{kingma2016improved}.

\subsection{MAF and RealNVP}

LINFA implements two widely used normalizing flow formulations, MAF~\cite{papamakarios2018masked} and RealNVP~\cite{dinh2016density}.
MAF belongs to the class of \emph{autoregressive} normalizing flows. Given the latent variable $\bm{z} = (z_{1},z_{2},\dots,z_{d})$, it assumes $p(z_i|z_{1},\dots,z_{i-1}) = \phi[(z_i - \mu_i) / e^{\alpha_i}]$, where $\phi$ is the standard normal distribution, $\mu_i = f_{\mu_i}(z_{1},\dots,z_{i-1})$, $\alpha_i = f_{\alpha_i}(z_{1},\dots,z_{i-1}),\,i=1,2,\dots,d$, and $f_{\mu_i}$ and $f_{\alpha_i}$ are masked autoencoder neural networks~\cite[MADE,][]{germain2015made}. 
In a MADE autoencoder the network connectivities are multiplied by Boolean masks so the input-output relation maintains a lower triangular structure, making the computation of the Jacobian determinant particularly simple. 
MAF transformations are then composed of multiple MADE layers, possibly interleaved by batch normalization layers~\cite{ioffe2015batch}, typically used to add stability during training and increase network accuracy~\cite{papamakarios2018masked}.

RealNVP is another widely used flow where, at each layer the first $d'$ variables are left unaltered while the remaining $d-d'$ are subject to an affine transformation of the form $\widehat{\bm{z}}_{d'+1:d} = \bm{z}_{d'+1:d}\,\odot\,e^{\bm{\alpha}} + \bm{\mu}$, where $\bm{\mu} = f_{\mu}(\bm{z}_{1:d'})$ and $\bm{\alpha} = f_{\alpha}(\bm{z}_{d'+1:d})$ are MADE autoencoders. 
In this context, MAF could be seen as a generalization of RealNVP by setting $\mu_i=\alpha_i=0$ for $i\leq d'$~\cite{papamakarios2018masked}.

\subsection{Normalizing flow with adaptive surrogate (NoFAS)}

LINFA is designed to accommodate black-box models $\bm{f}: \bm{\mathcal{Z}} \to \bm{\mathcal{X}}$ between the random inputs $\bm{z} = (z_1, z_2, \cdots, z_d)^T \in \bm{\mathcal{Z}}$ and the outputs $(x_1, x_2,\cdots,x_m)^T \in \bm{\mathcal{X}}$, and assumes $n$ observations $\bm x = \{\bm x_i\}_{i=1}^n \subset \bm{\mathcal{X}}$ to be available. 
Our goal is to infer $\bm z$ and to quantify its uncertainty given $\bm{x}$. 
We employ a variational Bayesian paradigm and sample from the posterior distribution $p(\bm z\vert \bm x)\propto \ell_{\bm z}(\bm x,\bm{f})\,p(\bm z)$, with prior $p(\bm z)$ via normalizing flows. 

This requires the evaluation of the gradient of the ELBO~\eqref{equ:ELBO} with respect to the NF parameters $\bm{\lambda}$, replacing $p(\bm x, \bm z_K)$ with $p(\bm x\vert\bm z_K)\,p(\bm z)$ $=\ell_{\bm z_K}(\bm{x},\bm{f})\,p(\bm z)$, and approximating the expectations with their MC estimates. 
However, the likelihood function needs to be evaluated at every MC realization, which can be costly if the model $\bm{f}(\bm{z})$ is computationally expensive. In addition, automatic differentiation through a legacy (e.g. physics-based) solver may be an impractical, time-consuming, or require the development of an adjoint solver.

Our solution is to replace the model $\bm{f}$ with a computationally inexpensive surrogate $\widehat{\bm{f}}: \bm{\mathcal{Z}} \times \bm{\mathcal{W}} \to \bm{\mathcal{X}}$ parameterized by the weigths $\bm{w} \in \bm{\mathcal{W}}$, whose derivatives can be obtained at a relatively low computational cost, but intrinsic bias in the selected surrogate formulation, a limited number of training examples, and locally optimal $\bm{w}$ can compromise the accuracy of $\widehat{\bm{f}}$.

To resolve these issues, LINFA implements NoFAS, which updates the surrogate model adaptively by smartly weighting the samples of $\bm{z}$ from NF thanks to a \emph{memory-aware} loss function.
Once a newly updated surrogate is obtained, the likelihood function is updated, leading to a new posterior distribution that will be approximated by VI-NF, producing, in turn, new samples for the next surrogate model update, and so on. 
Additional details can be found in~\cite{wang2022variational}.

\subsection{Adaptive Annealing}

Annealing is a technique to parametrically smooth a target density to improve sampling efficiency and accuracy during inference. 
In the discrete case, this is achieved by incrementing an \emph{inverse temperature} $t_{k}$ and setting $p_k(\bm{z},\bm{x}) = p^{t_k}(\bm{z},\bm{x}),\,\,\text{for } k=0,\dots,K$, where $0 < t_{0} < \cdots < t_{K} \le 1$.
The result of exponentiation produces a smooth unimodal distribution for a sufficiently small $t_0$, recovering the target density as $t_{k}$ approaches 1. In other words, annealing provides a continuous deformation from an easier to approximate unimodal distribution to a desired target density.

A linear annealing scheduler~\cite[see, e.g.,][]{rezende2015variational} with fixed temperature increments is often used in practice, where \mbox{$t_j=t_{0} + j (1-t_{0})/K$} for \mbox{$j=0,\ldots,K$} with constant increments  \mbox{$\epsilon = (1-t_{0})/K$}. 
Intuitively, small temperature changes are desirable to carefully explore the parameter spaces at the beginning of the annealing process, whereas larger changes can be taken as $t_{k}$ increases, after annealing has helped to capture important features of the target distribution (e.g., locating all the relevant modes).

The AdaAnn scheduler determines the increment $\epsilon_{k}$ that approximately produces a pre-defined change in the KL divergence between two distributions annealed at~$t_{k}$ and $t_{k+1}=t_{k}+\epsilon_{k}$, respectively.
Letting the KL divergence equal a constant $\tau^2/2$, where $\tau$ is referred to as the \emph{KL tolerance}, the step size $\epsilon_k$ becomes 
\begin{equation}\label{equ:adaann}
\epsilon_k = \tau/ \sqrt{\mathbb{V}_{p^{t_k}}[\log p(\bm z,\bm{x})]}. 
\end{equation}
The denominator is large when the support of the annealed distribution $p^{t_{k}}(\bm{z},\bm{x})$ is wider than the support of the target $p(\bm{z},\bm{x})$, and progressively reduces with increasing $t_{k}$.
Further detail on the derivation of the expression for $\epsilon_{k}$ can be found in~\cite{cobian2023adaann}.

\section{Detailed numerical benchmarks}\label{sec:detailed_benchmarks}

\subsection{Simple two-dimensional map with Gaussian likelihood}

A model $f:\mathbb{R}^{2}\to \mathbb{R}^{2}$ is chosen in this experiment having the closed-form expression
\begin{equation}
f(\bm z) = f(z_{1},z_{2}) = (z_1^3 / 10 + \exp(z_2 / 3), z_1^3 / 10 - \exp(z_2 / 3))^T.
\end{equation}
Observations $\bm{x}$ are generated as
\begin{equation}\label{eqn:exp1}
\bm{x} = \bm{x}^{*} + 0.05\,|\bm{x}^{*}|\,\odot\bm{x}_{0},
\end{equation}
where $\bm{x}_{0} \sim \mathcal{N}(0,\bm I_2)$ and $\odot$ is the Hadamard product. 
We set the \emph{true} model parameters at $\bm{z}^{*} = (3, 5)^T$, with output $\bm{x}^{*} = f(\bm z^{*})=(7.99, -2.59)^{T}$, and simulate 50 sets of observations from~\eqref{eqn:exp1}. The likelihood of $\bm z$ given $\bm{x}$ is assumed Gaussian and we adopt a noninformative uniform prior $p(\bm z)$.
We allocate a budget of $4\times4=16$ model solutions to the pre-grid and use the rest to adaptively calibrate $\widehat{f}$ using $2$ samples every $1000$ normalizing flow iterations.

\begin{figure}[!ht]
\centering
\includegraphics[scale=0.7]{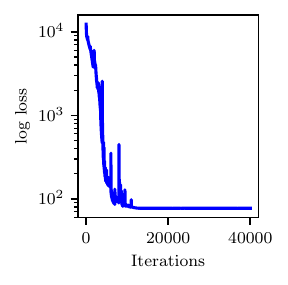}
\includegraphics[scale=0.75]{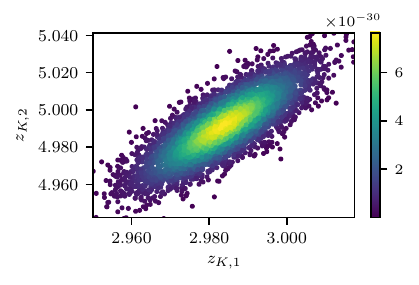}
\includegraphics[scale=0.7]{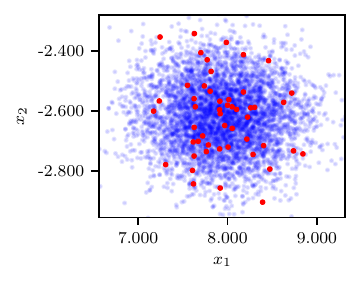}
\caption{Results from the trivial model. Loss profile (left), posterior samples (center), and posterior predictive distribution (right).}\label{fig:trivial}
\end{figure}

Results in terms of loss profile, variational approximation, and posterior predictive distribution are shown in Figure~\ref{fig:trivial}.

\subsection{High-dimensional example}

\begin{figure}[!ht]
\centering
\includegraphics[scale=0.7]{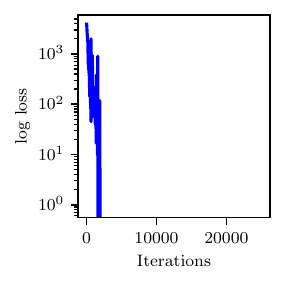}
\includegraphics[scale=0.7]{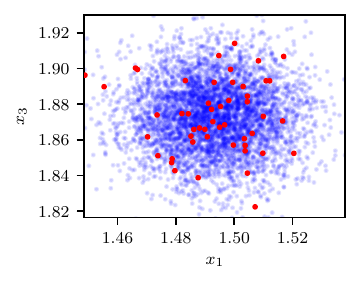}
\includegraphics[scale=0.7]{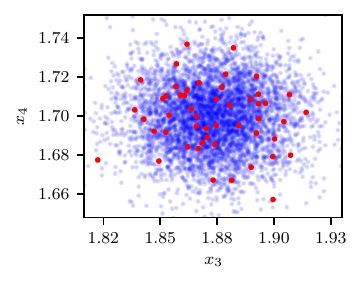}\\
\includegraphics[scale=0.7]{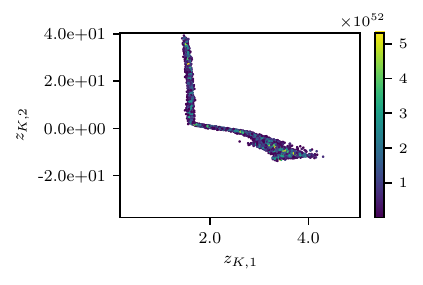}
\includegraphics[scale=0.7]{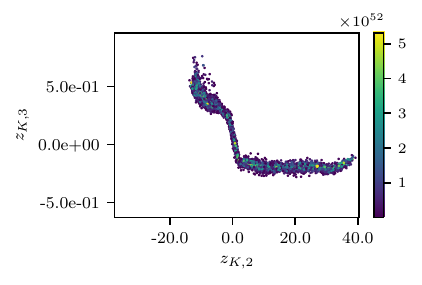}
\includegraphics[scale=0.7]{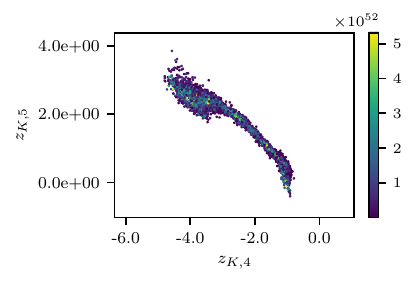}
\caption{Results from the high-dimensional model. Loss profile, posterior samples, and posterior predictive distribution.}\label{fig:highdim}
\end{figure}

We consider a map $f: \mathbb{R}^{5}\to\mathbb{R}^{4}$ expressed as
\begin{equation}
f(\bm{z}) = \bm{A}\,\bm{g}(e^{\bm{z}}),
\end{equation}
where $g_i(\bm{r}) = (2\cdot |2\,a_{i} - 1| + r_i) / (1 + r_i)$ with $r_i > 0$ for $i=1,\dots,5$ is the \emph{Sobol}  function~\cite{sobol2003theorems} and $\bm{A}$ is a $4\times5$ matrix. We also set
\begin{equation*}
\bm{a} = (0.084, 0.229, 0.913, 0.152, 0.826)^T \mbox{ and }\bm{A} = \frac{1}{\sqrt{2}}
\begin{pmatrix}
1 & 1 & 0 & 0 & 0\\
0 & 1 & 1 & 0 & 0\\
0 & 0 & 1 & 1 & 0\\
0 & 0 & 0 & 1 & 1\\
\end{pmatrix}.
\end{equation*}
The true parameter vector is set at $\bm{z}^{*} = (2.75,$ $-1.5, 0.25,$ $-2.5,$ $1.75)^T$. While the Sobol function is bijective and analytic, $f$ is over-parameterized and non identifiabile.
This is also confirmed by the fact that the curve segment $\gamma(t) = g^{-1}(g(\bm z^*) + \bm v\,t)\in Z$ gives the same model solution as $\bm{x}^{*} = f(\bm{z}^{*}) = f(\gamma(t)) \approx (1.4910,$ $1.6650,$ $1.8715,$ $1.7011)^T$ for $t \in (-0.0153, 0.0686]$, where $\bm v = (1,-1,1,-1,1)^T$. 
This is consistent with the one-dimensional null-space of the matrix $\bm A$.
We also generate synthetic observations from the Gaussian distribution
\begin{equation}
\bm{x} = \bm{x}^{*} + 0.01\cdot |\bm{x}^{*}| \odot \bm{x}_{0},\,\,\text{and}\,\,\bm{x}_{0} \sim \mathcal{N}(0,\bm I_5).
\end{equation}
Results are shown in Figure~\ref{fig:highdim}.

\subsection{Two-element Windkessel Model}

\begin{figure}[!ht]
\centering
\includegraphics[scale=0.7]{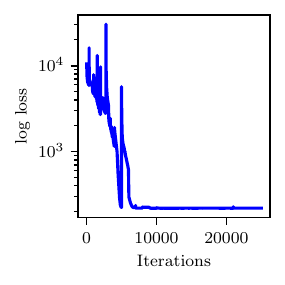}
\includegraphics[scale=0.75]{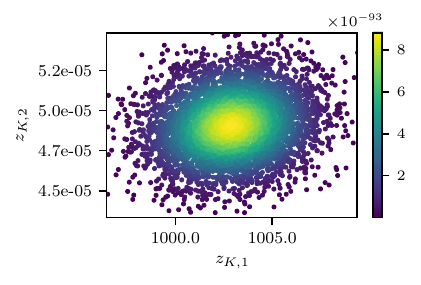}
\includegraphics[scale=0.7]{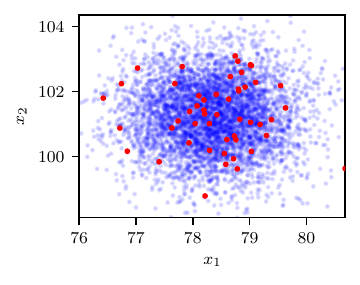}
\caption{Results from the RC model. Loss profile (left), posterior samples (center) for R and C, and the posterior predictive distribution for $P_{p,\text{min}}$ and $P_{p,\text{max}}$ (right, $P_{p,\text{avg}}$ not shown).}\label{fig:rc_res}
\end{figure}

The two-element Windkessel model (often referred to as the \emph{RC} model) is the simplest representation of the human systemic circulation and requires two parameters, i.e., a resistance $R \in [100, 1500]$ Barye$\cdot$ s/ml and a capacitance $C \in [1\times 10^{-5}, 1 \times 10^{-2}]$ ml/Barye. 
We provide a periodic time history of the aortic flow (see~\cite{wang2022variational} for additional details) and use the RC model to predict the time history of the proximal pressure $P_{p}(t)$, specifically its maximum, minimum, and average values over a typical heart cycle, while assuming the distal resistance $P_{d}(t)$ as a constant in time, equal to 55 mmHg. 
In our experiment, we set the true resistance and capacitance as $z_{K,1}^{*}=R^{*} = 1000$ Barye$\cdot$ s/ml and $z_{K,2}^{*}=C^{*} = 5\times 10^{-5}$ ml/Barye, and determine $P_{p}(t)$ from a RK4 numerical solution of the following algebraic-differential system
\begin{equation}\label{equ:RC}
Q_{d} = \frac{P_{p}-P_{d}}{R},\quad \frac{d P_{p}}{d t} = \frac{Q_{p} - Q_{d}}{C},
\end{equation}
where $Q_{p}$ is the flow entering the RC system and $Q_{d}$ is the distal flow.
Synthetic observations are generated by adding Gaussian noise to the true model solution $\bm{x}^{*}=(x^{*}_{1},x^{*}_{2},x^{*}_{3})=(P_{p,\text{min}},$ $P_{p,\text{max}},$ $P_{p,\text{avg}})= (78.28, 101.12,  85.75)$, i.e., $\bm{x}$ follows a multivariate Gaussian distribution with mean $\bm{x}^{*}$ and a diagonal covariance matrix with entries $0.05\,x_{i}^{*}$, where $i=1,2,3$ corresponds to the maximum, minimum, and average pressures, respectively. 
The aim is to quantify the uncertainty in the RC model parameters given 50 repeated pressure measurements. We imposed a non-informative prior on $R$ and $C$. Results are shown in Figure~\ref{fig:rc_res}.

\subsection{Three-element Wndkessel Circulatory Model (NoFAS)}

The three-parameter Windkessel or \emph{RCR} model is characterized by proximal and distal resistance parameters $R_{p}, R_{d} \in [100, 1500]$ Barye$\cdot$s/ml, and one capacitance parameter $C \in [1\times 10^{-5}, 1\times 10^{-2}]$ ml/Barye.  
This model is not identifiable. The average distal pressure is only affected by the total system resistance, i.e. the sum $R_{p}+R_{d}$, leading to a negative correlation between these two parameters. Thus, an increment in the proximal resistance is compensated by a reduction in the distal resistance (so the average distal pressure remains the same) which, in turn, reduces the friction encountered by the flow exiting the capacitor. An increase in the value of $C$ is finally needed to restore the average, minimum and maximum pressure. This leads to a positive correlation between $C$ and $R_{d}$.

The output consists of the maximum, minimum, and average values of the proximal pressure $P_{p}(t)$, i.e., $(P_{p,\text{min}}, P_{p,\text{max}}, P_{p,\text{avg}})$ over one heart cycle.
The true parameters are $z^{*}_{K,1} = R^{*}_{p} = 1000$ Barye$\cdot$s/ml, $z^{*}_{K,2}=R^{*}_{d} = 1000$ Barye$\cdot$s/ml, and $C^{*} = 5\times 10^{-5}$ ml/Barye. The proximal pressure is computed from the solution of the algebraic-differential system
\begin{equation}
Q_{p} = \frac{P_{p} - P_{c}}{R_{p}},\quad Q_{d} = \frac{P_{c}-P_{d}}{R_{d}},\quad \frac{d\, P_{c}}{d\,t} = \frac{Q_{p}-Q_{d}}{C},
\end{equation}
where the distal pressure is set to $P_{d}=55$ mmHg.
Synthetic observations are generated from $N(\bm\mu, \bm\Sigma)$, where $\mu=(f_{1}(\bm{z}^{*}),f_{2}(\bm{z}^{*}),f_{3}(\bm{z}^{*}))^T = (P_{p,\text{min}}, P_{p,\text{max}}, P_{p,\text{ave}})^T = (100.96,$ $148.02,$ $ 116.50)^T$ and $\bm\Sigma$ is a diagonal matrix with entries $(5.05, 7.40, 5.83)^T$. The budgeted number of true model solutions is $216$; the fixed surrogate model is evaluated on a $6\times 6\times 6 = 216$ pre-grid while the adaptive surrogate is evaluated with a pre-grid of size $4\times 4\times 4 = 64$ and the other 152 evaluations are adaptively selected.

\begin{figure}[!ht]
\centering
\includegraphics[scale=0.7]{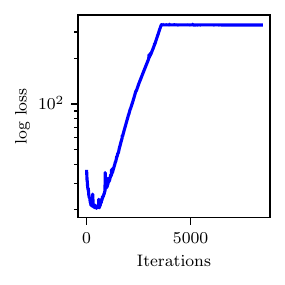}
\includegraphics[scale=0.7]{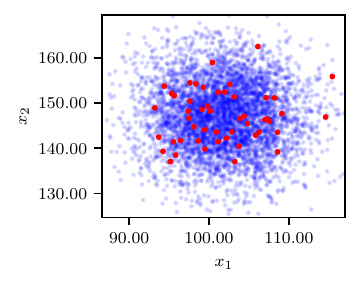}
\includegraphics[scale=0.7]{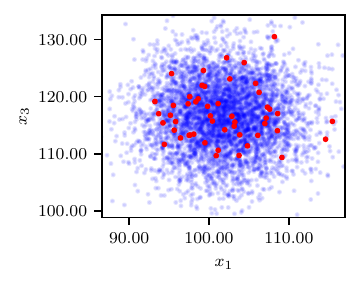}\\
\includegraphics[scale=0.65]{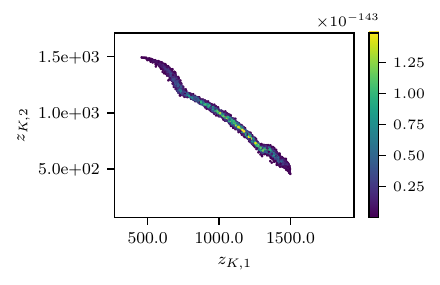}
\includegraphics[scale=0.65]{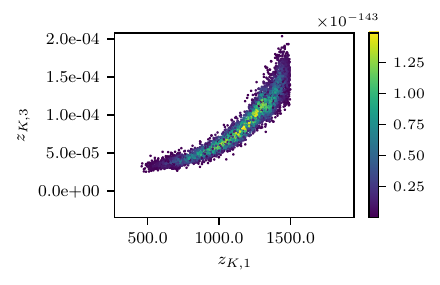}
\includegraphics[scale=0.65]{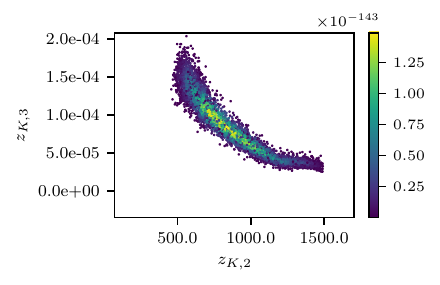}
\caption{Results from the RCR model. Loss profile (left), posterior predictive distribution (center), and posterior samples (right).}\label{fig:rcr_res}
\end{figure}

This example also demonstrates how NoFAS can be combined with annealing for improved convergence. The results in Figure ~\ref{fig:rcr_res} are generated using the AdaAnn adaptive annealing scheduler with intial inverse temperature $t_{0}=0.05$, KL tolerance $\tau=0.01$ and a batch size of 100 samples. The number of parameter updates is set to 500, 5000 and 5 for $t_{0}$, $t_{1}$ and $t_{0}<t<t_{1}$, respectively and 1000 Monte Carlo realizations are used to evaluate the denominator in equation~\eqref{equ:adaann}. The posterior samples capture well the nonlinear correlation among the parameters and generate a fairly accurate posterior predictive distribution that overlaps with the observations. Additional details can be found in~\cite{wang2022variational,cobian2023adaann}.

\subsection{Friedman 1 model (AdaAnn)}

We consider a modified version of the Friedman 1 dataset~\cite{friedman1991multivariate} to examine the performance of  our adaptive annealing scheduler in a high-dimensional context. 
According to the original model in~\cite{friedman1991multivariate}, the data are generated as
\begin{equation}\label{eqn:friedman1}
\textstyle y_i = \mu_i(\boldsymbol{\beta})+ \epsilon_i, \mbox{ where }
\mu_i(\boldsymbol{\beta})=\beta_1\text{sin}(\pi x_{i,1}x_{i,2})+ \beta_2(x_{i,3}-\beta_3)^2+\sum_{j=4}^{10}\beta_jx_{i,j}, 
\end{equation}
where $\epsilon_i\sim\mathcal{N}(0,1)$. 
We made a slight modification to the model in~\eqref{eqn:friedman1} as
\begin{equation} \label{eqn:friedman1_modified}
\mu_i(\boldsymbol{\beta}) = \textstyle \beta_1\text{sin}(\pi x_{i,1}x_{i,2})+ \beta_2^2(x_{i,3}-\beta_3)^2+\sum_{j=4}^{10}\beta_jx_{i,j},
\end{equation}
and set the true parameter combination to $\boldsymbol{\beta}=(\beta_1,\ldots,\beta_{10})=(10,\pm \sqrt{20}, 0.5, 10, 5, 0, 0, 0, 0, 0)$. Note that both~\eqref{eqn:friedman1} and \eqref{eqn:friedman1_modified} contain linear, nonlinear, and interaction terms of the input variables $X_1$ to $X_{10}$, five of which ($X_6$ to $X_{10}$) are irrelevant to $Y$. Each $X$ is drawn independently from $\mathcal{U}(0,1)$. We used R package \texttt{tgp} \cite{gramacy2007tgp} to generate a Friedman~1 dataset with a sample size of $n$=1000.
We impose a non-informative uniform prior $p(\boldsymbol{\beta})$ and, unlike the original modal, we now expect a bimodal posterior distribution of $\boldsymbol{\beta}$. Results in terms of marginal statistics and their convergence for the mode with positive $z_{K,2}$ are illustrated in Table~\ref{table:Friedman_bimodal_stats} and Figure~\ref{fig:adaann_res}.

\vspace{10pt}

\begin{minipage}{\textwidth}
  \begin{minipage}[b]{0.4\textwidth}
    \centering
    \resizebox{.8\textwidth}{!}{%
    \begin{tabular}[2in]{l c c c c}
    \toprule
    \textbf{True} & \multicolumn{2}{c}{\textbf{Mode 1}}\\
    \textbf{Value} & Post. Mean & Post. SD\\
    \midrule
    $\beta_1 = 10$   & 10.0285 & 0.1000\\
    $\beta_2 = \pm \sqrt{20}$ & 4.2187 & 0.1719\\
    $\beta_3 = 0.5$  & 0.4854 & 0.0004\\
    $\beta_4 = 10$   & 10.0987 & 0.0491\\
    $\beta_5 = 5$    & 5.0182 & 0.1142\\
    $\beta_6 = 0$    & 0.1113 & 0.0785\\
    $\beta_7 = 0$    & 0.0707 & 0.0043\\
    $\beta_8 = 0$    & -0.1315 & 0.1008\\
    $\beta_9 = 0$    & 0.0976 & 0.0387\\
    $\beta_{10} = 0$ & 0.1192 & 0.0463\\
    \bottomrule
    \end{tabular}}  
    \captionof{table}{Posterior mean and standard deviation for positive mode in the modified Friedman test case.}\label{table:Friedman_bimodal_stats}    
  \end{minipage}
\hfill
\begin{minipage}[b]{0.58\textwidth}
\centering
\includegraphics[width=0.4\textwidth]{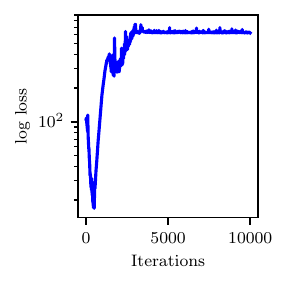}
\includegraphics[width=0.58\textwidth]{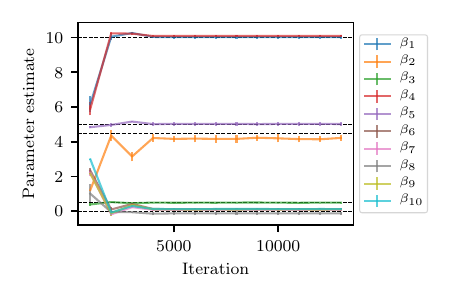}
\captionof{figure}{Loss profile (left) and posterior marginal statistics (right) for positive mode in the modified Friedman test case.}\label{fig:adaann_res}
\end{minipage}
\end{minipage}
%

\section{Hyperparameters in LINFA}\label{sec:hyper}

This section contains the list of all hyperparameters in the library, their default values, and a description of the functionalities they control.
General hyperparameters are listed in Table~\ref{tab:par_general}, those related to the optimization process in Table~\ref{tab:par_optimizers}, and to the output folder and files in Table~\ref{tab:par_output}. Hyperparameters for the proposed NoFAS and AdaAnn approaches are listed in Table~\ref{tab:surr_optimizers} and~\ref{tab:adaann}, respectively. 
Finally, a hyperparameter used to select the hardware device is described in Table~\ref{tab:par_device}.

\begin{table}[H]
\centering
\caption{Output parameters}\label{tab:par_output}
\begin{tabular}{p{4cm} p{2cm} p{8cm}} 
\toprule
{\bf Option} & {\bf Type} & {\bf Description}\\
\midrule
\emph{\texttt{output\_dir}} & string & Name of output folder where results files are written.\\
\emph{\texttt{log\_file}} & string & Name of the log file which stores the iteration number, annealing temperature, and value of the loss function at each iteration.\\
\emph{\texttt{seed}} & int & Seed for random number generator.\\
\bottomrule
\end{tabular}
\end{table}

\begin{table}[H]
\centering
\caption{Surrogate model parameters (NoFAS)}\label{tab:surr_optimizers}
\begin{tabular}{p{4cm} p{2cm} p{8cm}} 
\toprule
{\bf Option} & {\bf Type} & {\bf Description}\\
\midrule
\emph{\texttt{n\_sample}} & int & Batch size used to generate results after \texttt{save\_interval} iterations.\\
\emph{\texttt{calibrate\_interval}} & int & Number of NF iteration between successive updates of the surrogate model (default \emph{\texttt{1000}}).\\
\emph{\texttt{budget}} & int & Maximum allowable number of true model evaluations.\\
\emph{\texttt{surr\_pre\_it}} & int & Number of pre-training iterations for surrogate model (default \emph{\texttt{40000}}).\\
\emph{\texttt{surr\_upd\_it}} & int & Number of iterations for the surrogate model update (default \emph{\texttt{6000}}).\\
\emph{\texttt{surr\_folder}} & string & Folder where the surrogate model is stored (default \emph{\texttt{'./'}}).\\
\emph{\texttt{use\_new\_surr}} & bool & Start by pre-training a new surrogate and ignore existing surrogates (default \emph{\texttt{True}}).\\
\emph{\texttt{store\_surr\_interval}} & int & Save interval for surrogate model (\emph{\texttt{None}} for no save, default \emph{\texttt{None}}).\\
\bottomrule
\end{tabular}
\end{table}

\begin{table}[H]
\centering
\caption{Device parameters}\label{tab:par_device}
\begin{tabular}{p{4cm} p{2cm} p{8cm}} 
\toprule
{\bf Option} & {\bf Type} & {\bf Description}\\
\midrule
\emph{\texttt{no\_cuda}} & bool & Do not use GPU acceleration.\\
\bottomrule
\end{tabular}
\end{table}

\begin{table}[H]
\centering
\caption{Optimizer and learning rate parameters}\label{tab:par_optimizers}
\begin{tabular}{p{4cm} p{2cm} p{8cm}} 
\toprule
{\bf Option} & {\bf Type} & {\bf Description}\\
\midrule
\emph{\texttt{optimizer}} & string & Type of SGD optimizer (default \emph{\texttt{'Adam'}}).\\
\emph{\texttt{lr}} & float & Learning rate (default \emph{\texttt{0.003}}).\\
\emph{\texttt{lr\_decay}} & float & Learning rate decay (default \emph{\texttt{0.9999}}).\\
\emph{\texttt{lr\_scheduler}} & string & Type of learning rate scheduler (\emph{\texttt{'StepLR'}} or 
\emph{\texttt{'ExponentialLR'}}).\\
\emph{\texttt{lr\_step}} & int & Number of steps before learning rate reduction for the step scheduler.\\
\emph{\texttt{log\_interval}} & int & Number of iterations between successive loss printouts (default \emph{\texttt{10}}).\\
\bottomrule
\end{tabular}
\end{table}

\begin{table}[H]
\centering
\caption{General parameters}\label{tab:par_general}
\begin{tabular}{p{4cm} p{2cm} p{8cm}} 
\toprule
{\bf Option} & {\bf Type} & {\bf Description}\\
\midrule
\emph{\texttt{name}} & str & Name of the experiment.\\
\emph{\texttt{flow\_type}} & str & type of normalizing flow (\emph{\texttt{'maf'}},\emph{\texttt{'realnvp'}}).\\ 
\emph{\texttt{n\_blocks}} & int & Number of normalizing flow layers (default \emph{\texttt{5}}).\\
\emph{\texttt{hidden\_size}} & int & Number of neurons in MADE and RealNVP hidden layers (default \emph{\texttt{100}}).\\
\emph{\texttt{n\_hidden}} & int & Number of hidden layers in MADE (default 1).\\
\emph{\texttt{activation\_fn}} & str & Activation function for MADE network used by MAF (default \emph{\texttt{'relu'}}).\\
\emph{\texttt{input\_order}} & str & Input order for MADE mask creation (\emph{\texttt{'sequential'}} or \emph{\texttt{'random'}}, default \emph{\texttt{'sequential'}}).\\
\emph{\texttt{batch\_norm\_order}} & bool & Adds batchnorm layer after each MAF or RealNVP layer (default \emph{\texttt{True}}).\\
\emph{\texttt{save\_interval}} & int & How often to save results from the normalizing flow iterations. Saved results include posterior samples, loss profile, samples from the posterior predictive distribution, observations, and marginal statistics.\\
\emph{\texttt{input\_size}} & int & Input dimensionality (default \emph{\texttt{2}}).\\
\emph{\texttt{batch\_size}} & int & Number of samples from the basic distribution generated at each iteration (default \emph{\texttt{100}}).\\
\emph{\texttt{true\_data\_num}} & int & Number of additional true model evaluations at each surrogate model update (default \emph{\texttt{2}}).\\
\emph{\texttt{n\_iter}} & int & Total number of NF iterations (default \emph{\texttt{25001}}).\\
\bottomrule
\end{tabular}
\end{table}

\begin{table}[H]
\centering
\caption{Parameters for the adaptive annealing scheduler (AdaAnn)}\label{tab:adaann}
\begin{tabular}{p{4cm} p{2cm} p{8cm}} 
\toprule
{\bf Option} & {\bf Type} & {\bf Description}\\
\midrule
\emph{\texttt{annealing}} & bool & Flag to activate the annealing scheduler. If this is \emph{\texttt{False}}, the target posterior distribution is left unchanged during the iterations.\\
\emph{\texttt{scheduler}} & string & Type of annealing scheduler (\emph{\texttt{'AdaAnn'}} or \emph{\texttt{'fixed'}}, default \emph{\texttt{'AdaAnn'}}).\\
\emph{\texttt{tol}} & float & KL tolerance. It is kept constant during inference and used in the numerator of equation~\eqref{equ:adaann}.\\
\emph{\texttt{t0}} & float & Initial inverse temperature.\\
\emph{\texttt{N}} & int & Number of batch samples during annealing.\\
\emph{\texttt{N\_1}} & int & Number of batch samples at $t=1$.\\
\emph{\texttt{T\_0}} & int & Number of initial parameter updates at $t_0$.\\
\emph{\texttt{T}} & int & Number of parameter updates after each temperature update. During such updates the temperature is kept fixed.\\
\emph{\texttt{T\_1}} & int & Number of parameter updates at $t=1$\\
\emph{\texttt{M}} & int & Number of Monte Carlo samples used to evaluate the denominator in equation~\eqref{equ:adaann}.\\
\bottomrule
\end{tabular}
\end{table}

\vskip 0.2in
\bibliographystyle{unsrtnat}
\bibliography{linfa.bib}

\end{document}